\documentclass[sigconf, screen, nonacm]{acmart}
\usepackage{graphicx}
\usepackage{adjustbox}
\usepackage{amsmath}
\usepackage{balance}
\usepackage{soul} 
\usepackage{color, xcolor} 
\usepackage{makecell}
\usepackage{multirow}
\usepackage{enumitem}
\usepackage{enumerate}
\usepackage{colortbl}
\definecolor{1}{RGB}{255,146,146}
\definecolor{2}{RGB}{255,204,153}
\definecolor{3}{RGB}{255,255,153}
\definecolor{check}{RGB}{0,153,76}
\definecolor{fork}{RGB}{233,75,75}
\usepackage{array}
\newcommand{\PreserveBackslash}[1]{\let\temp=\\#1\let\\=\temp}
\newcolumntype{C}[1]{>{\PreserveBackslash\centering}p{#1}}
\newcolumntype{R}[1]{>{\PreserveBackslash\raggedleft}p{#1}}
\newcolumntype{L}[1]{>{\PreserveBackslash\raggedright}p{#1}}
\usepackage{pifont}

\makeatletter
\def\@ACM@checkaffil{% Only warnings
    \if@ACM@instpresent\else
    \ClassWarningNoLine{\@classname}{No institution present for an affiliation}%
    \fi
    \if@ACM@citypresent\else
    \ClassWarningNoLine{\@classname}{No city present for an affiliation}%
    \fi
    \if@ACM@countrypresent\else
        \ClassWarningNoLine{\@classname}{No country present for an affiliation}%
    \fi
}
\makeatother

\AtBeginDocument{%
  }
    
\begin{document}    

%\copyrightyear{2025} 
%\acmYear{2025} 
%\setcopyright{acmlicensed}
%\acmConference[MM '25]{Proceedings of the 33nd ACM International Conference on Multimedia}{October 27-31}{Dublin, Ireland}

\settopmatter{printacmref=false}

\title{SuperGS: Consistent and Detailed 3D Super-Resolution Scene Reconstruction via Gaussian Splatting}

%%%%%%%%% AUTHORS - PLEASE UPDATE
\author{Shiyun Xie}
% \orcid{0000-0001-5921-4060}
\affiliation{%
  \institution{School of Artificial Intelligence, Beihang University}
  \city{Beijing}
  \country{China}
}
\email{xieshiyun@buaa.edu.cn}

\author{Zhiru Wang}
% \orcid{0009-0004-3794-2699}
\affiliation{%
  %\department{}
  \institution{Sino-French Engineer School, Beihang University}
  \city{Beijing}
  \country{China}
}
\email{19241085@buaa.edu.cn}

\author{Yinghao Zhu}
\affiliation{%
  %\department{}
  \institution{School of Artificial Intelligence, Beihang University}
  \city{Beijing}
  \country{China}
}
\email{zhuyinghao@buaa.edu.cn}

\author{Xu Wang}
\affiliation{%
  %\department{}
  \institution{School of Artificial Intelligence, Beihang University}
  \city{Beijing}
  \country{China}
}
\email{21421012@buaa.edu.cn}

\author{Chengwei Pan}
\authornote{Corresponding authors}
% \orcid{0000-0003-0497-7903}
\affiliation{%
  %\department{School of Artificial Intellegence}
  \institution{School of Artificial Intelligence, Beihang University}
}
\affiliation{%
  % \institution{Zhongguancun Laboratory}
  \city{Beijing}
  \country{China}
}
\email{pancw@buaa.edu.cn}

\author{Xiwang Dong}
% \orcid{0000-0003-0497-7903}
\affiliation{%
  \institution{Institute of Unmanned System, Beihang University}
}
\affiliation{%
  % \institution{Zhongguancun Laboratory}
  \city{Beijing}
  \country{China}
}
\email{xwdong@buaa.edu.cn}

\begin{abstract}

Recently, 3D Gaussian Splatting (3DGS) has excelled in novel view synthesis (NVS) with its real-time rendering capabilities and superior quality. However, it encounters challenges for high-resolution novel view synthesis (HRNVS) due to the coarse nature of primitives derived from low-resolution input views. To address this issue, we propose SuperGS, an expansion of Scaffold-GS designed with a two-stage coarse-to-fine training framework. In the low-resolution stage, we introduce a latent feature field to represent the low-resolution scene, which serves as both the initialization and foundational information for super-resolution optimization. In the high-resolution stage, we propose a multi-view consistent densification strategy that backprojects high-resolution depth maps based on error maps and employs a multi-view voting mechanism, mitigating ambiguities caused by multi-view inconsistencies in the pseudo labels provided by 2D prior models while avoiding Gaussian redundancy. Furthermore, we model uncertainty through variational feature learning and use it to guide further scene representation refinement and adjust the supervisory effect of pseudo-labels, ensuring consistent and detailed scene reconstruction. Extensive experiments demonstrate that SuperGS outperforms state-of-the-art HRNVS methods on both forward-facing and 360-degree datasets. 
%Code is available at \url{https://anonymous.4open.science/r/SuperGS}.
\end{abstract}




\begin{teaserfigure}
\centering
  \includegraphics[scale=.3]{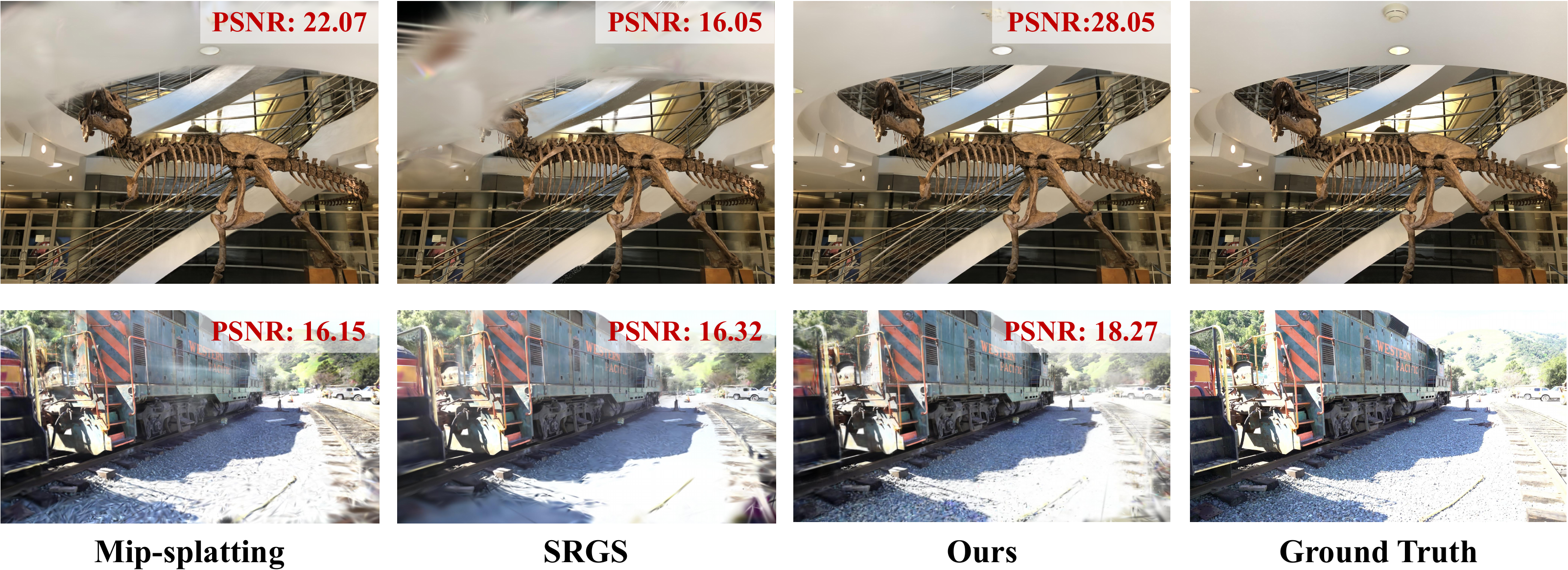}
  \caption{\textit{Comparisons in High-resolution Novel View Synthesis.} SuperGS demonstrates more detailed and high-fidelity results compared to existing GS-based methods, delivering consistent advantages in extremely challenging scenes.}
\end{teaserfigure}

\maketitle

\section{Introduction}

Novel view synthesis (NVS) is crucial for applications like AR/VR, autonomous navigation, and 3D content creation in the field of computer vision and graphics~\cite{deng2022fov, tonderski2024neurad, poole2022dreamfusion}. Traditional methods using meshes and points often sacrifice rendering quality for higher speed~\cite{munkberg2022extracting, botsch2005high, yifan2019differentiable}. Conversely, Neural Radiance Fields (NeRF)~\cite{mildenhall2021nerf, barron2021mip, barron2022mip} have significantly advanced these tasks by implicitly modeling scene geometry and radiance, but their high computational demands restrict their application in real-time scenarios~\cite{muller2022instant, fridovich2022plenoxels, Chen2022ECCV}. Recently emerging as a promising approach, 3D Gaussian Splatting (3DGS)~\cite{kerbl20233d} leverages 3D Gaussian primitives and a differentiable rasterization process for real-time, high-quality rendering. This technique eliminates the need for extensive ray sampling and incorporates clone-and-split strategies to improve spatial coverage and enhance scene representation. 

However, when handling high-resolution novel view synthesis (HRNVS), vanilla 3DGS suffers from significant performance degradation. Unlike NeRF-style models, which can sample the color and opacity of any point due to their continuous implicit scene representations, 3DGS cannot directly upsample Gaussian primitives. Existing methods typically use 2D priors to guide high-resolution scene optimization. The straightforward approach is to leverage image super-resolution models to generate pseudo-labels~\cite{feng2024srgs}, but processing images individually introduces multi-view inconsistencies. Some works propose using video super-resolution (VSR) models to ensure consistency. For example, Supergaussian~\cite{shen2024supergaussian} utilizes low-resolution reconstruction results to render continuous trajectories as videos, but this amplifies errors from the low-resolution reconstruction and offers no mechanism for correction. Sequence Matters~\cite{ko2024sequence} leverages the similarity between input views to obtain "video-like" sequences, but this requires dense inputs with small angular differences, limiting its practical applications. Considering the greater availability and abundance of image data, current image models possess superior feature extraction and detail restoration capabilities, while video models comparatively demand higher computational resources. Therefore, we opt to utilize a pretrained single image super-resolution (SISR) model to generate high-resolution input views as pseudo labels to guide detail enhancement.

To mitigate pseudo-label errors and achieve high-quality high-resolution reconstruction, we introduce SuperGS. This method employs a coarse-to-fine framework to ensure high fidelity, while proposing improved densification and uncertainty-aware learning strategies that simultaneously preserve reconstruction details and consistency. Due to Scaffold-GS's excellent novel view synthesis performance and low memory requirements, which are significant advantages when representing high-resolution scenes, we select it as our backbone. Scaffold-GS incorporates implicit representations into 3D Gaussian Splatting, enabling us to learn scene representations at the latent feature level rather than through explicit color and shape parameters. We further enhance Scaffold-GS by introducing a latent feature field to represent the low-resolution scene, serving as both the initialization and foundation for super-resolution optimization.

% To mitigate pseudo-label errors and achieve high-fidelity reconstruction, we propose a two-stage coarse-to-fine framework. In this framework, we enhance Scaffold-GS by introducing a latent feature field to represent the low-resolution scene, serving as both the initialization and foundation for super-resolution optimization. This approach not only provides an initial coarse feature for high-resolution learning but also ensures that additional details remain consistent with the core scene information, thereby guaranteeing high fidelity while reducing the impact of pseudo-label errors.

Furthermore, due to multi-view inconsistencies, gradient-based densification tends to generate excessive Gaussians to fit each training view, resulting in an overabundance of Gaussians and performance degradation in novel view synthesis. To address this, we introduce a multi-view voting densification strategy, replacing the original anchor growth policy of Scaffold-GS. This method generates candidate points based on pixel-wise loss and upsampled high-resolution depth map backprojection, then determines anchor growth through a multi-view voting mechanism, ensuring multi-view consistency while avoiding Gaussian redundancy. Additionally, we model uncertainty by learning variational features, which help identify regions where the model lacks confidence. Subsequently, we leverage this uncertainty to further enhance the scene representation and adjust the supervision from pseudo-labels. 

To evaluate the effectiveness of our method, we conducted extensive experiments on the challenging real-world and LLFF datasets. Experimental results validate that our method consistently outperforms existing methods, achieving consistent and detailed 3D super-resolution scene reconstruction in complex environments.

In summary, the main contributions of our approach are as follows :
\begin{itemize}[leftmargin=*]
    \item \textit{Insightfully}, we introduce SuperGS, an expansion of Gaussian Splatting for 3D scene super-resolution. SuperGS incorporates 2D priors to guide detail enhancement while ensuring multi-view consistency and high-fidelity reconstruction.
    
    \item \textit{Methodologically}, we incorporate a latent feature field throughout our two-stage coarse-to-fine framework, which enables effective information propagation. Additionally, the multi-view consistent densification and uncertainty-aware learning strategy further ensure 3D consistent and detailed super-resolution scene reconstruction.
    
    \item \textit{Experimentally}, extensive experiments demonstrate that SuperGS consistently surpasses state-of-the-art methods on both forward-facing datasets and 360-degree datasets.
\end{itemize}

\begin{figure*}[!ht]
\centering
\includegraphics[width=1.0\linewidth]{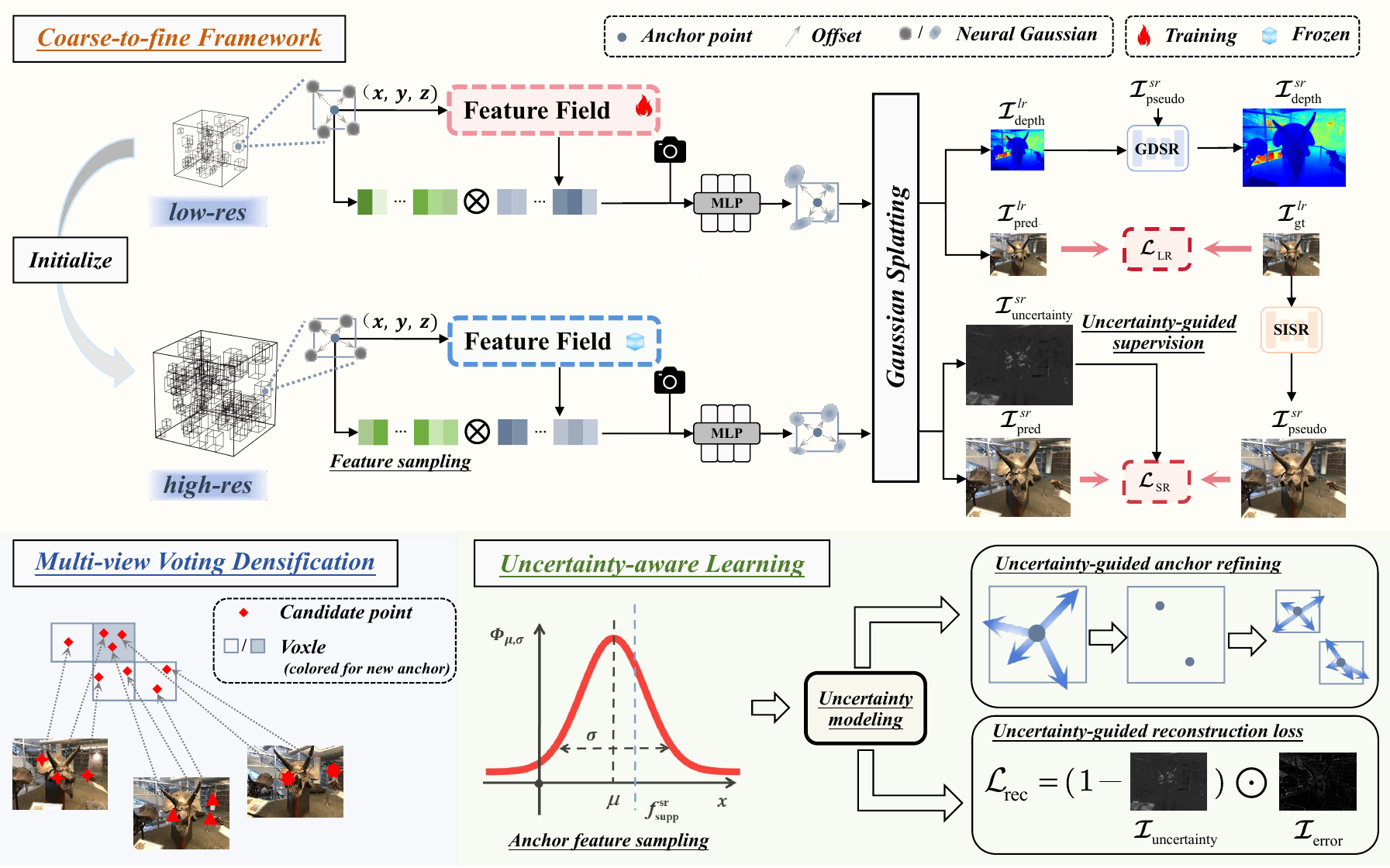}
\caption{\textit{Framework of our proposed SuperGS.} (a) We propose a two-stage coarse-to-fine framework. We enhance Scaffold-GS by introducing a latent feature field to represent the low-resolution scene, which serves as both initialization and foundational information for super-resolution optimization. (b) In the high-resolution stage, we propose a multi-view consistent densification strategy that replaces the original gradient-based densification, avoiding overfitting and Gaussian redundancy. (c) We model anchor uncertainty through learning variational features, which is further used to guide scene representation refinement and reconstruction loss computation.}
\label{fig:framework}
% \vspace{-5mm}
\end{figure*}
\section{Related Work}
% \label{sec:formatting}

\subsection{Novel View Synthesis}

Novel view synthesis (NVS) is a complex task in computer vision, aimed at generating new viewpoint images from a set of posed photos. Neural radiance fields (NeRF)~\cite{mildenhall2021nerf} have excelled in NVS, encoding scene details into a MLP for volume rendering. However, NeRF's limitations in detail capture and efficiency led to the development of Mip-NeRF~\cite{barron2021mip} and InstantNGP~\cite{muller2022instant}, which enhance both aspects. Subsequent research~\cite{liu2020neural, hedman2021baking, lin2022efficient} has focused on finding a balance between rendering quality and efficiency, a persistently challenging task.

The recent development of 3D Gaussian Splatting (3DGS)~\cite{kerbl20233d} offers a potential solution to overcome the limitations of NeRF. By converting scene point clouds into anisotropic Gaussian primitives following Structure from Motion (SFM)~\cite{schonberger2016structure}, 3DGS uses these primitives for explicit scene representation and employs a differentiable rasterizer for real-time, high-quality rendering. Subsequent 3DGS improvements have either enhanced reconstruction quality through anti-aliasing~\cite{yu2024mip, yan2024multi} and depth regulation~\cite{li2024dngaussian, turkulainen2024dn, chung2024depth}, or focused on model compression~\cite{lee2024compact, chen2024hac, niedermayr2024compressed}. Additionally, some works~\cite{lu2024scaffold, ren2024octree} have incorporated implicit representations into 3DGS, improving performance while significantly reducing memory requirements. Various adaptations have also tailored 3DGS to specific scenarios such as sparse view reconstruction~\cite{xiong2024sparsegs, zhu2023FSGS, fan2024instantsplat}, spectacular scene reconstruction~\cite{yang2024spec, wang2024specgaussian}, and dynamic scenes~\cite{duan20244d, yang2023real, huang2024endo}.

\subsection{3D Scene Super-Resolution}
High-resolution novel view synthesis (HRNVS) aims to generate high-resolution novel views from low-resolution inputs. NeRF-SR~\cite{wang2022nerf} addresses this by sampling multiple rays per pixel and applying regularization at the sub-pixel level. However, it requires ground truth high-resolution images as input, which are often unavailable in practical scenarios. RefSR-NeRF~\cite{huang2023refsr} faces similar limitations. FastSR-NeRF~\cite{lin2024fastsr} introduces Random Patch Sampling to accelerate the process, which is crucial for high-resolution scenes. Furthermore, pretrained 2D models provide valuable priors for 3D super-resolution tasks. To ensure multi-view consistency, \cite{yoon2023cross} employs cross-guided optimization, while DiSR-NeRF\cite{lee2024disr} utilizes iterative 3D synchronization and enhances image details through Renoised Score Distillation. 

% NeRF-based methods still 受限于渲染速度，特别是在high-resolution的场景下。Some works 开始探索利用3DGS来做HRNVS task。SRGS~\cite{feng2024srgs} uses high-resolution inputs from SISR as pseudo labels and introduces sub-pixel constraints as regularization. 由于image model单独处理每张输入视角图像，会带来multi-view nconsistencies. SuperGaussian~\cite{shen2025supergaussian} attempts to address this by sampling a smooth trajectory from a 3D scene to generate video, and then using a pretrained video super-resolution (VSR) model to produce supervision labels. 然而，由于渲染的连续视频来自低分辨率重建结果rather than ground truth，上采样后会放大重建错误而没法再修正。同样地，这个问题同样出现在GaussianSR里，which uses 低分辨率渲染图像作为2D super-resolution model StableSR的condition。SequenceMatters~\cite{} address this by 通过度量input vies的相似度来生成'video-like' sequences，但是it requires 密集且拍摄角度相差较小的输入图像，which限制了实际应用。并且由于images are much easier to obtain than videos, existing 2D image models still offer superior feature representation 和细节恢复能力，并且video model也需要更多的计算资源. Therefore, we still consider relying on a SISR model as the 2D prior. we 提出了coarse-to-fine framework 来保证高保真度，并提出了改进的densification strategy和uncertainty-aware learning来保证多视角一致性。

To ensure real-time rendering, recent work has begun exploring 3D Gaussian Splatting (3DGS) for HRNVS tasks. SRGS~\cite{feng2024srgs} uses high-resolution outputs from SISR as pseudo labels and introduces sub-pixel constraints as regularization. However, since image models process each input view independently, this introduces multi-view inconsistencies. SuperGaussian~\cite{shen2024supergaussian} attempts to address this by sampling smooth trajectories from 3D scenes to generate videos, then using pretrained video super-resolution (VSR) models for supervision. However, since the rendered continuous video comes from low-resolution reconstruction rather than ground truth, upsampling amplifies reconstruction errors without correcting them. GaussianSR faces similar issues, using low-resolution rendered images as conditions for the 2D super-resolution model StableSR~\cite{wang2024exploiting}. Sequence Matters~\cite{ko2024sequence} addresses this by measuring input view similarity to generate "video-like" sequences, but requires dense inputs with small angular differences, limiting practical applications. Since images are more readily available than videos, existing 2D image models still offer superior feature representation and detail recovery, while video models require more computational resources. Therefore, we still rely on SISR models as 2D priors. We propose a coarse-to-fine framework to ensure high fidelity, along with improved densification strategies and uncertainty-aware learning to guarantee multi-view consistency.
\section{Preliminaries}
% \label{sec:formatting}

\subsection{3DGS}

3D Gaussian Splatting (3DGS)~\cite{kerbl20233d} models scenes using anisotropic 3D Gaussians and renders images by rasterizing their projected 2D counterparts. Each 3D Gaussian $\mathcal{G}(x)$ is parameterized by a center position $\mu \in \mathbb{R}^3$ and a covariance matrix $\Sigma \in \mathbb{R}^{3\times3}$:
\begin{equation}
    \mathcal{G}(x) = e^{-\frac{1}{2}(x-\mu)^T \Sigma^{-1}(x-\mu)}
\end{equation}

The covariance matrix $\Sigma$ is parameterized through a scaling matrix $S \in \mathbb{R}^3$ and rotation matrix $R \in \mathbb{R}^{3\times3}$ with $\Sigma = RSS^TR^T$. Each Gaussian has associated opacity $\sigma \in \mathbb{R}^1$ and color feature $F \in \mathbb{R}^C$. The rendering process employs a tile-based rasterizer that sorts 3D Gaussians in front-to-back order and projects them onto the image plane as 2D Gaussians. The color at each pixel is computed through alpha blending:
\begin{equation}
   \hat{C} = \sum_{i \in N} c_i \alpha_i T_i
\end{equation}
where$N$ represents the number of sorted 2D Gaussians associated with that pixel, and $T$ denotes the transmittance as $\prod_{j=1}^{i-1} (1 - \alpha_j)$. 

The scene is initialized with Gaussian means and colors derived from SfM~\cite{schonberger2016structure}. During optimization, Gaussian parameters are updated through gradient descent across multiple rendering iterations to best match the training images. At fixed intervals, the algorithm splits, clones, and prunes Gaussians based on their opacity, screen-space size, and mean gradient magnitude, respectively.

\subsection{Scaffold-GS}

Scaffold-GS~\cite{lu2024scaffold} enhances 3D-GS by introducing a more efficient and structured primitive management system through anchors. Each anchor is associated with features that describe the local structure and emits multiple neural Gaussians:
\begin{equation}
    \{\mu_0, \ldots, \mu_{k-1}\} = X + \{O_0, \ldots, O_{k-1}\} \cdot S
\end{equation}
where $X$ is the anchor position, ${\mu_i}$ represents positions of neural Gaussians, ${O_i}$ are predicted offsets, and $S$ is a scaling factor. Properties like opacities, rotations, and colors are decoded from anchor features $f$ through MLPs:
\begin{equation}
    \{\alpha_0, \ldots, \alpha_{k-1}\} = \text{F}_{\alpha}(f, \delta_{vc}, \vec{\mathbf{d}}_{vc})
\end{equation}
where $\delta_{vc}$ and $\vec{\mathbf{d}}_{vc}$ are relative viewing distance and direction between the camera and the anchor point.

For anchor points refinement, Scaffold-GS employs both growing and pruning operations. The growing operation adds new anchors in voxels where the gradient of neural Gaussians exceeds a threshold, allowing better representation in texture-less areas. The system quantizes space into multi-resolution voxels for adaptive refinement. Conversely, the pruning operation removes anchors that consistently produce low-opacity Gaussians, maintaining computational efficiency while preserving visual quality.
\section{Method}

\label{sec:Method}

In this work, we introduce a Super-Resolution Gaussian Splatting method (SuperGS) for high-resolution novel view synthesis (HRNVS) from low-resolution input views, utilizing a two-stage coarse-to-fine training framework. In coarse stage training, we obtain a rough scene representation using low-resolution ground-truth images, while in fine stage training, we enhance the details by leveraging high-resolution pseudo-labels generated from a pre-trained Single Image Super-Resolution (SISR) model~\cite{liang2021swinir}. The two-stage process is described as follows.

\subsection{Coarse-to-fine Framework}

\subsubsection{Hash-grid latent feature field}

% The framework of SuperGS is shown in Figure~\ref{fig:framework}. Ours method 基于Scaffold-GS，首先利用低分辨率真值图像Scaffold-GS得到低分辨率重建结果。不同的是，我们将anchor feature $f$分为两部分表示: $f_{hash} \otimes f_{anc}$，which is further decoded to neural gaussians parameters like opacity, scale and so on. 其中 $f_{hash}$ 由latent feature field得到，which用于学习场景的低分辨率场景信息，并并用于后续高分辨率阶段的foundation feature. 由于$f_{hash}$ 由特征场根据anchor位置插值得到，会有信息损失。因此我们仍在每个anchor上保留了一个低维的可学习feature用于信息补充。

The framework of SuperGS is shown in Figure~\ref{fig:framework}. Our method is based on Scaffold-GS, first utilizing low-resolution ground truth images to obtain low-resolution reconstruction representation through Scaffold-GS. The key difference is that we decompose the anchor feature $f$ into two parts: $f_{\text{field}} \otimes f^{\text{lr}}_{\text{supp}}$, which is subsequently decoded to neural Gaussians attributes.

\begin{figure}[!ht]
\centering
\includegraphics[width=0.7\linewidth]{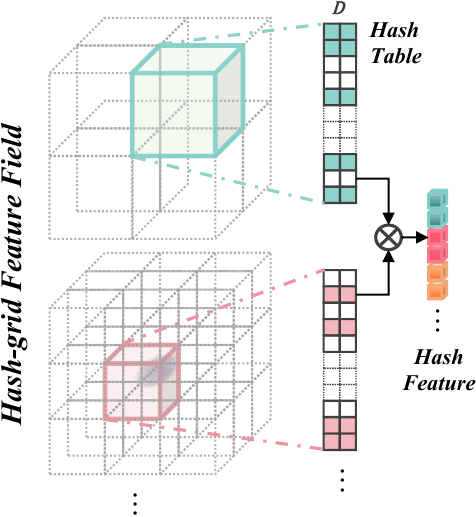}
\caption{\textit{Illustration of Feature Field.} For a specific anchor, we extract and interpolate features from hash tables using its coordinates, with the concatenation of features from $L$ resolution levels forming the field feature $f_{\text{field}}$ of this anchor.}
\label{fig:feature_field}
% \vspace{-3mm}
\end{figure}

Within this approach, $f_{\text{field}}$ is derived from a latent feature field, which learns low-resolution scene information and serves as the foundational feature for subsequent high-resolution stage. Drawing inspiration from InstantNGP~\cite{muller2022instant}, we adopt a similar strategy by storing view-independent features in multi-resolution grids. This approach allows for efficient access to anchor features at arbitrary positions by indexing through hash tables followed by linear interpolation. 
% As illustrated in Figure~\ref{fig:feature_field}, 
For each anchor, we retrieve learnable feature vectors stored in grids with varying resolutions and perform linear interpolation to derive its feature representation. Notably, different levels of grids correspond to distinct hash tables. Specifically, there are $L$ levels of grid resolutions, and the feature vector interpolated for the $i$-th anchor at the $l$-th level is denoted as $f_{\text{field}}^{(l)}$. From this, we construct its multi-resolution feature vector representation as follows:
\begin{equation}
    f_{\text{field}} = f_{\text{field}}^{(1)} \otimes f_{\text{field}}^{(2)} \otimes ... \otimes f_{\text{field}}^{(L)}
\end{equation}
where $\otimes$ denotes the concatenation operation.

Similarly, to enable the grid feature field to extend to unbounded scenes, we reference the approach used in Mip-NeRF360~\cite{barron2021mip}, which normalizes the coordinates of the anchors into a contracted space. The contraction is formally described as follows:
\begin{equation}
    \mathrm{contract}(X_i) = \left\{\begin{array}{ll}
X_i, & ||X_i|| \le 1 \\
(2-\frac{1}{||X_i||})(\frac{X_i}{||X_i||}), & ||X_i|| > 1
    \end{array}\right.
\end{equation}
where $X_i \in \mathbb{R}^3$ represents the position coordinate of $i$-th anchor.

Since $f_{\text{field}}$ is interpolated from the feature field based on anchor positions,  inevitably resulting in some information loss, we retain a low-dimensional learnable feature $f^{\text{lr}}_{\text{supp}}$ at each anchor to supplement the missing information. Then, followed by Scaffold-GS, the neural Gaussians attributes $\alpha$, $q$, $s$ and $c$ are decoded by anchor feature. For example, opacity values of neural Gaussians spawned from an anchor point are given by:

\begin{equation}
    \{\alpha_0, ..., \alpha_{k-1}\} = F_{\alpha}(f_{\text{field}}, f^{\text{lr}}_{\text{supp}}, \delta_{vc}, \vec{\mathbf{d}}_{vc}),
\end{equation}

\subsubsection{Variational anchor feature}
\label{sec:variational}

% During the fine training stage, we freeze the parameters of the feature field, and 将其通过位置索引插值得到的feature $f_{\text{field}}$ 作为 base feature. Additionally, we still attach a learnable feature vector $f^{sr}_{\text{field}}$ to each anchor to enhance the super-resolution details, which is initialized by low-resolution feature $f_{\text{supp}}$. 
% 由于我们利用2D pretrained SISR model 来得到高分辨率伪标签，其多视角不一致性会导致模型学习产生歧义。因此，我们通过将$f^{sr}_{\text{field}}$建模为variational feature，从而建模每个anchor的不确定性，which is used to subsequent anchor refinement and loss computation~\ref{sec:Uncertainty}. 具体地，we assume that $f^{sr}_{\text{field}}$ follow a normal distribution $\mathcal{N}(f_\mu, \exp(f_\sigma))$, learning both the mean and variance, where the exponential operation ensures the variance is non-negative. To achieve forward propagation and ensure gradients can backpropagate, we use the reparameterization trick~\cite{kingma2013auto} for residual feature sampling:

% \begin{equation}
%     f^{sr}_{\text{field}}=f_{\mu_i}+{\epsilon}\cdot f_{\sigma_i}, \quad \epsilon \sim \mathcal{N} \mathcal(0, 1)
% \end{equation}

% where $f_{\mu_i}$ is initialized by $f^{lr}_{\text{field}}$, and $f_{\sigma_i}$ 被初始化为很小的值。

During the fine training stage, we freeze the parameters of the feature field, and use the interpolated feature $f_{\text{field}}$ obtained through position indexing as the base feature. Additionally, we still attach a learnable feature vector $f^{\text{sr}}_{\text{supp}}$ to each anchor to enhance high-resolution details, which is initialized by the low-resolution feature $f^{\text{lr}}_{\text{supp}}$.

Since we utilize a 2D pretrained SISR model to obtain high-resolution pseudo-labels, multi-view inconsistencies can lead to ambiguity during model training. Therefore, we model $f^{\text{sr}}_{\text{field}}$ as a variational feature for  representing the uncertainty of each anchor, which subsequently guides anchor refinement and loss computation as described in Section~\ref{sec:Uncertainty}. Specifically, we assume that $f^{\text{sr}}_{\text{supp}}$ follows a normal distribution $\mathcal{N}(f_\mu, \exp(f_\sigma))$, learning both the mean and variance, where the exponential operation ensures the variance is non-negative. To achieve forward propagation and ensure that gradients can backpropagate, we use the reparameterization trick~\cite{kingma2013auto} for feature sampling:

\begin{equation}
f^{\text{sr}}_{\text{supp}}=f_{\mu}+{\epsilon}\cdot \exp(f_\sigma), \quad \epsilon \sim \mathcal{N}(0, 1)
\end{equation}
where $f_{\mu}$ is initialized by $f^{\text{lr}}_{\text{supp}}$, and $f_{\sigma}$ is initialized with very small values.

Unlike previous two-stage methods ~\cite{hu2024gaussiansr, shen2024supergaussian, feng2024zs}, which only utilize low-resolution reconstruction as initialization for high-resolution reconstruction, our method establishes a latent feature field that represents the low-resolution 3D scene, serving as the foundational feature for the high-resolution representation. This approach not only provides an initial coarse feature for high-resolution anchors in the subsequent fine-stage training but also ensures that additional details remain consistent with the core scene information. This guarantees high fidelity in the high-resolution scene reconstruction while mitigating the impact of errors from pseudo-labels.

\subsection{Multi-view Consistent Densification}

% 实验表明，在fine training stage 阶段，使用3DGS和Scaffold-GS中基于梯度的adaptive densification strategy会损失性能。这可能源于伪标签的多视角不一致性，导致其需要更多的高斯来拟合每个视角的信息，导致新视角合成出现较大偏差。另外，一些方法~\cite{}也指出基于梯度的densification strategy会导致高纹理区域欠拟合，而简单地降低阈值又会导致大量的高斯冗余。因此，我们结合了像素级的误差和深度图，并利用多视角投票机制来决定anchor的增长。

\begin{figure}[!ht]
\centering
\includegraphics[width=1.0\linewidth]{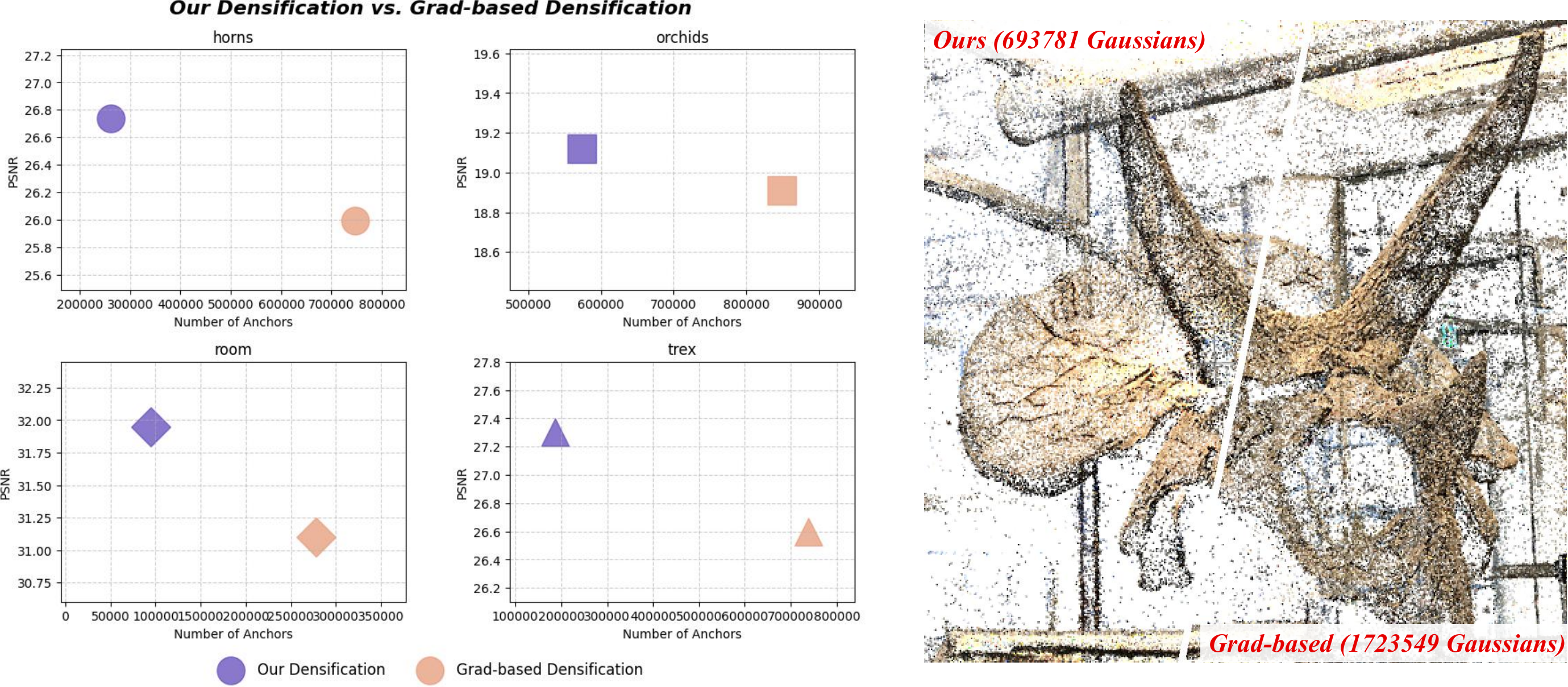}
% 我们的方法在使用更少anchor数量的同时保证了更好的重建质量。这不仅减少了内存需求，也避免了过拟合。
\caption{\textit{Comparison of Ours Densification and Gradient-based Densification Startegy.} Our method achieves better reconstruction quality with fewer anchor points, reducing memory requirements while preventing overfitting.}
\label{fig:densify_comparison}
% \vspace{-3mm}
\end{figure}

Experiments demonstrate that during the fine training stage, using the gradient-based adaptive densification strategy from 3DGS and Scaffold-GS can degrade performance, as shown in Figure~\ref{fig:densify_comparison}. This may stem from multi-view inconsistencies in pseudo-labels, requiring more Gaussians to fit texture from each viewpoint, thus causing significant deviations in novel view synthesis. Additionally, some methods~\cite{zhang2024pixel, rota2024revising} have indicated that gradient-based densification strategies can lead to underfitting in highly textured regions, while simply lowering thresholds results in excessive Gaussian redundancy. Therefore, we combine pixel-wise errors with depth map and utilize a multi-view voting mechanism to determine anchor growth.

\subsubsection{Depth map rendering and super-resolution}
\label{sec: depth render}

% 深度图提供了丰富的几何结构信息，而利用单目深度估计模型得到的深度图会存在与实际训练场景尺度不一致的问题~/cite{DNGaussian}。因此，我们首先渲染低分辨率场景得到低分辨率深度图，再利用 pretrained guided depth map super-resolution model~\cite{SGNet} to obatain high-resolution depth map, which 是利用高分辨率rgb图像来指导低分辨率深度图上采样。基于此，我们得到的高分辨率深度图不仅尺度匹配，也和rgb伪标签信息一致。

% !!!!!!!!!增加：插值上采样depth map的问题

% 对于低分辨率深度图渲染，目前常用alpha blending 渲染得到深度图，或者选取the largest contributing $w_i=T_i\alpha_i$ 来 represents the depth of that pixel. 前者的像素深度值会收到光线上所有高斯的影响，这会得到一个更为平滑的深度图但lack of 精确的3D结构。而后者只考虑权重最高的高斯，这在透明场景中会存在问题. 参考~\cite{SparseGS}, 低分辨率深度图渲染时我们apply softmax-scaling来优化原始的alpha blending的权重:
% \begin{equation}
% d^{lr}_{x,y}^{\text{softmax}} = \log \left( \frac{\sum_{i=1}^{N} w_i e^{\beta w_i} d_i}{\sum_{i=1}^{N} w_i e^{\beta w_i}} \right)
% \end{equation}
% 指数项的使用使得depth accumulating 增大了高不透明度高斯的贡献，减少了低权重点的噪声影响。其中$\beta$用于调节softmax温度，越大则权重差异越明显。

Depth maps provide rich geometric structural information, though depth maps obtained from monocular depth estimation models often have scale inconsistencies with the actual training scene~\cite{turkulainen2024dn}. To address this, we first render low-resolution depth maps, then employ a pretrained guided depth map super-resolution (GDSR) model~\cite{wang2024sgnet} to obtain high-resolution depth maps, which uses high-resolution RGB images to guide the upsampling of low-resolution depth maps. This approach yields high-resolution depth maps that are not only scale-matched but also consistent with geometric structure information in the pseudo-labels.

For low-resolution depth map rendering, common approaches include alpha blending or selecting the largest contributing weight $w_i=T_i\alpha_i$ to represent the depth of each pixel. The former results in pixel depth values influenced by all Gaussians along the ray, producing smoother depth maps but lacking precise 3D structure. The latter only considers the highest-weighted Gaussian, which can be problematic in transparent scenes. Following the approach in~\cite{xiong2024sparsegs}, we apply softmax-scaling to optimize the original alpha blending weights during low-resolution depth map rendering:

\begin{equation}
d_{u,v}^{lr} = \log \left( \frac{\sum_{i=1}^{N} w_i e^{\beta w_i} d_i}{\sum_{i=1}^{N} w_i e^{\beta w_i}} \right)
\end{equation}
The exponential term enhances the contribution of high-opacity Gaussians while reducing noise influence from low-weight Gaussians. The parameter $\beta$ adjusts the softmax temperature, with higher values creating more pronounced weight differentiation.

\subsubsection{Error-based depth projection and multi-view voting mechanism}

\begin{figure}[!ht]
\centering
\includegraphics[width=0.9\linewidth]{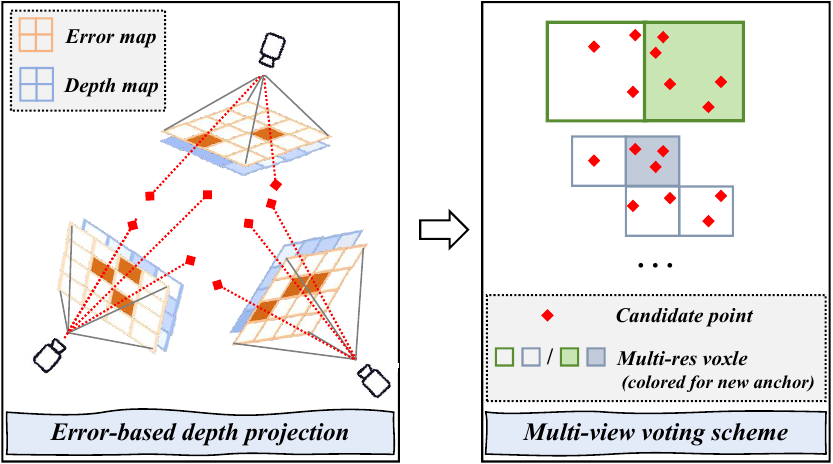}
% We introduce a multi-vew voting densification strategy replacing orin anchor growing policy. 首先我们基于pixel loss 反投影对应像素的深度得到candidate Gaussian positions, then new anchors are added while the votes in a voxel exceed a threshold.
\caption{\textit{Illustration of Densification Strategy.} We introduce a multi-view voting densification strategy that replaces the original anchor growing policy. First, we generate candidate Gaussian positions by back-projecting the corresponding pixels with depth map based on pixel-wise loss. Subsequently, new anchors are added when the accumulated vote count within a voxel exceeds a predetermined threshold.}
\label{fig:densification}
% \vspace{-3mm}
\end{figure}

The multi-view stereo (MVS) methods~\cite{chen2019point, yao2018mvsnet, zhang2023geomvsnet} utilize depth projection to point clouds for 3D reconstruction, but their depth maps typically come from sensors and can only reconstruct areas with high confidence. In contrast, we have already obtained high-quality pseudo high-resolution depth maps from Section~\ref{sec: depth render}, which contain excellent geometric prior information. However, pixel-level depth projection leads to over-reconstruction and brings unacceptable memory requirements for GS-based methods. Based on these considerations, we propose a multi-view consistent densification mechanism.

First, we define under-reconstruction regions as areas where the loss between rendered images and high-resolution pseudo-labels is large, and backproject the depth from these regions to obtain candidate Gaussian positions. Specifically, given the camera intrinsics $K_k$ and extrinsics $[R_k, t_k]$ for the $k$-th viewpoint, we can project a specific pixel coordinate $p_{k,i}=[u_{k,i}, v_{k,i}]^T$ into 3D space using its corresponding depth value $d_{k,i}$, resulting in the 3D coordinate $X_{k,i}$:
\begin{equation}
X_{k,i} = R_k^T K_k^{-1} \begin{bmatrix} u_{k,i} d_{k,i} \ v_{k,i} d_{k,i} \ d_{k,i} \end{bmatrix} + t_k
\end{equation}

Retaining all projected candidate points would result in tremendous memory usage and lead to overfitting to training views, with errors becoming more pronounced when pseudo-labels are inconsistent. Therefore, we propose a multi-view voting mechanism to densify anchors in under-reconstructed regions. While Scaffold-GS defines each voxel as a unit and uses the sum of gradients of neural Gaussians within each voxel as a condition for anchor growing, we instead use the number of candidate points in each voxel as a condition for anchor densification. We treat candidate points as votes from various training viewpoints in the corresponding region and add an anchor to voxels where the point count exceeds a threshold. Similarly, we employ a multi-resolution voxel grid to allow new anchors to be added at different granularities, with regions containing more candidate points receiving anchors at higher levels. 
% The multi-view consistent densification strategy pseudocode is shown in~\ref{code:densification}.

\subsection{Uncertainty-aware Learning}
\label{sec:Uncertainty}

% 我们利用2D pretrained SR model 得到的伪标签作为高分辨率场景学习的监督信号，which 在大量的2D图像上进行训练，能够提供很好的先验信息。然而single image super-resolution(SISR) model 单独处理每张低分辨率图像，会存在多视角不一致性。

% % 由于SISR模型得到的伪标签存在多视角不一致性，不一致区域会给场景学习造成干扰。As a remedy, we propose to model the supplementary anchor feature $f^{sr}_{\text{supp}}$ as a Gaussian distribution instead of a single feature vector, as described in Sec~\ref{sec:variational}. The predicted variance can serve as the reflection of uncertainty of the anchor. Formally, we define the uncertainty $u_i$ of $i$-th anchor using the magnitude of its feature variance:
% \begin{equation}
%     u_i=||\exp(f_\sigma)||_2
% \end{equation}

\subsubsection{Uncertainty modeling}

We leverage pseudo-labels generated by SISR models as supervision, though this approach introduces multi-view inconsistencies. To address this, we model the supplementary anchor feature $f^{sr}_{\text{supp}}$ as a Gaussian distribution rather than a single feature vector, as described in Section~\ref{sec:variational}. The predicted variance serves as an uncertainty indicator for each anchor. Formally, we define the uncertainty $u_i$ of the $i$-th anchor using the magnitude of its feature variance:
\begin{equation}
    u_i=||\exp(f_\sigma)||_2
\end{equation}

This uncertainty estimation allows us to identify problematic regions and implement targeted refinement strategies. Note that for the 360-degree datasets, we jointly train with three viewpoints per iteration as in MVGS~\cite{du2024mvgs}, which helps improve uncertainty modeling while accelerating convergence.

\subsubsection{Uncertainty-guided anchor refining}

% Intuitively, high uncertainty 反映了模型对该区域的重建结果is not confident. 因此, we propose an uncertainty-guided anchor refining strategy. Specifically, 对于不确定度超过阈值的anchor，我们将其细化为两个下一个level，即更小voxel的anchor。具体操作为，假设我们选定的anchor在第$l$个level的voxel内，我们便在该voxel随机采样两个位置. 然后利用$l+1$个level的voxel size来voxelize the points:
% \begin{equation}
%     \mathbf{X}_{\text{new}} = \left\{ \left[ \frac{\mathbf{X}_{\text{sample}}}{\epsilon^{l+1}} \right] \right\} \cdot \epsilon^{l+1}
% \end{equation}
% % where$\mathbf{X}_{\text{sample}$ 表示在原voxel内随机采样点的位置，$\mathbf{X}_{\text{new}}$ 表示细化后的voxel center, 也是新的anchor的坐标。其中两个相邻level的voxel size关系为:
% \begin{equation}
%     \epsilon^{(l+1)} = \epsilon^{(l)}/4
% \end{equation}

% 需要注意的是，我们需要确保两个采样点有一定距离，使其不会落在同一个voxel内，这可以通过使采样点与原voxel中心的偏移offset相反来实现。另外，如果细化后的voxel内已有anchor，则不再增加anchor。This approach is 某种程度上和3DGS的Gaussian Splitting想法类似，都利用更多的高斯来细化该区域的场景表达。需要指出，在anchor densification以及anchor refining阶段，我们都利用了Scaffold-GS的opacity-based anchor pruning 策略，确保anchor不会过于冗余。

Intuitively, high uncertainty values indicate areas where the model lacks confidence in its reconstruction. Therefore, we propose an uncertainty-guided anchor refining strategy. Specifically, anchors with uncertainty exceeding a threshold will be refined by splitting into two extra anchors, which are then added to corresponding voxels at the next level. Given an anchor located within a voxel at level $l$, we randomly sample two positions within this voxel and voxelize these points using the level $l+1$ voxel size:
\begin{equation}
    \mathbf{X}_{\text{new}} = \left\{ \left[ \frac{\mathbf{X}_{\text{sample}}}{\epsilon^{(l+1)}} \right] \right\} \cdot \epsilon^{(l+1)}
\end{equation}
where $\mathbf{X}_{\text{sample}}$ represents the randomly sampled positions within the original voxel, and $\mathbf{X}_{\text{new}}$ denotes the refined voxel centers, which serve as coordinates for the new anchors. The relationship between voxel sizes at adjacent levels is defined as:
\begin{equation}
    \epsilon^{(l+1)} = \epsilon^{(l)}/4
\end{equation}

Importantly, we ensure the two sampled points maintain sufficient distance to prevent them from falling within the same voxel, achieved by generating sampling offsets in opposite directions from the original voxel center. Additionally, if a refined voxel already contains an anchor, no new anchor is added. This approach shares conceptual similarities with Gaussian Splitting scheme in 3DGS~\cite{kerbl20233d}, as both utilize more Gaussians to refine scene representation in under-optimized regions. During both anchor densification and anchor refining stages, we employ opacity-based anchor pruning strategy of Scaffold-GS to prevent excessive anchor redundancy.

\subsubsection{Uncertainty-guided supervision}

% 另一方面，伪标签不一致的区域也使模型学习产生歧义，从而带来较大的不确定性。我们希望减弱这些区域的监督作用，从而减缓不一致性带来的影响。首先，我们需要得到每个视角的uncertainty map. 在NeRF相关方法中~/cite{}, 通常通过多次采样并多次渲染图像，利用像素的方差大小作为不确定性表征。但这种方式极其耗时。Thanks to Gaussian Splatting的显示表达，我们可以直接将不确定性通过alpha blending的方式渲染。给定第$k$个视角，the formula for 第$p$个pixel的uncertainty $U_{k,p}$ is:
% \begin{equation}
%     U_{k,p} = \sum_{i \in N} u_i a_i \prod_{j=1}^{i-1} (1 - a_j)
% \end{equation}
% 其中，我们将neural Gaussian的uncertainty取值为其归属的anchor的uncertainty for convenience. Then, we perform uncertainty-guided reconstruction loss insdead of 原来的L1 loss:
% \begin{equation}
%     \mathcal{L}_{\text{rec}} = 
%     ||(1-U(u,v))\odot (\mathcal{I}_{\text{pred}} - \mathcal{I}_{\text{pseudo}})||_1
% \end{equation}

% 由此，伪标签中higher confidence的区域对参数更新有较大作用，leading to more accurate representation of the scene.

The inconsistencies in pseudo-labels create ambiguities during model learning, resulting in increased uncertainty. We aim to reduce the influence of these inconsistent regions by weakening their supervisory role. First, we generate an uncertainty map for each view. Unlike NeRF-based methods ~\cite{shen2021stochastic, shen2022conditional} that typically estimate uncertainty map through multiple sampling and rendering iterations, a highly time-consuming process, we directly render uncertainties of Gaussians via alpha blending. For the $p$-th pixel in the $k$-th view, the uncertainty $U_{k,p}$ is calculated as:
\begin{equation}
    U_{k,p} = \sum_{i \in N} u_i \alpha_i \prod_{j=1}^{i-1} (1 - \alpha_j)
\end{equation}
where, for convenience, each neural Gaussian's uncertainty is assigned the value of its associated anchor's uncertainty. We then implement an uncertainty-guided reconstruction loss instead of the standard $\mathcal{L}_1$ loss:
\begin{equation}
    \mathcal{L}_{\text{rec}} =
    ||(1-\sigma(U(u,v)))\odot (\mathcal{I}_{\text{pred}} - \mathcal{I}_{\text{pseudo}})||_1
\end{equation}
where $\sigma(\cdot)$ represents sigmoid function.

This approach ensures that regions with higher confidence in the pseudo-labels exert greater influence on parameter updates, resulting in more accurate scene representation.

\begin{table*}[!ht]
        \footnotesize
        \centering
        \caption{\textit{Quantitative comparison for HRNVS ($\times 4$) on real-world datasets.} Mip-NeRF360: $1/8\rightarrow 1/2$, Deep Blending: $1/4\rightarrow 1/1$, Tansk\&Temples: $480\times270\rightarrow 1920\times1080$.}
        \label{tab:real_world}
    \begin{tabular}{L{2cm}|C{0.8cm}C{0.8cm}C{0.8cm}C{0.8cm}|C{0.8cm}C{0.8cm}C{0.8cm}C{0.8cm}|C{0.8cm}C{0.8cm}C{0.8cm}C{0.8cm}} % {c|ccc|ccc|ccc}
\toprule
Dataset     & \multicolumn{4}{c|}{Mip-NeRF360} & \multicolumn{4}{c|}{Deep Blending} & \multicolumn{4}{c}{Tanks\&Temples}   \\
Method\&Metric      & PSNR$\uparrow$   & SSIM$\uparrow$  & LPIPS$\downarrow$  &FID$\downarrow$  & PSNR$\uparrow$   & SSIM$\uparrow$  & LPIPS$\downarrow$ &FID$\downarrow$ & PSNR$\uparrow$   & SSIM$\uparrow$  & LPIPS$\downarrow$ &FID$\downarrow$ \\ \midrule
3DGS            &20.72 &0.617 &0.396 & 68.45
                &26.71 &0.839 &0.324 & 81.81
                &18.32 &0.612 &0.415 & 35.18\\

Scaffold-GS     &26.15 &0.739 &0.316 & 78.25
                &28.36 &0.872 &0.306 & 69.55
                &20.50 &0.667 &0.396 & 35.54 \\

Mip-splatting   &26.42 &0.756 &0.294 & 37.73
                &28.85 &0.884 &0.282 & 59.44
                &20.65 &0.672 &0.385 & 30.25 \\
                
GaussianSR      &25.60 &0.663 &0.368 & -
                &28.28 &0.873 &0.307 & -
                &- &- &- &-\\
                
SRGS            &26.88 &\textbf{0.767} &\textbf{0.286} & 46.37
                &29.40 &0.894 &0.272 & 60.96
                &20.70 &0.687 &0.379 & 34.58\\            
                \midrule
              
\textbf{Ours}   &\textbf{27.28} &\textbf{0.767} &0.290 &\textbf{34.26}  
                &\textbf{30.05} &\textbf{0.901} &\textbf{0.271} &\textbf{53.47}
                &\textbf{21.31} &\textbf{0.693} &\textbf{0.371} &\textbf{28.76}  \\

\bottomrule
\end{tabular}
% \vspace{-4mm}
\end{table*}

\subsection{Losses Design}

% 和Scaffold-GS相同，我们同样a使用了SSIM loss $\mathcal{L}_{\text{SSIM}}$ and volume regularization loss $\mathcal{L}_{\text{vol}}$. 和大多数super-resolution 方法~\cite{NeRF-SR,SRGS,GaussianSR}不同，我们没有利用低分辨率真值图像再对高分辨率渲染图像进行监督约束，这样会削弱其细节重建的能力。为了提高保真度，我们下采样了高分辨率渲染图像，并将其和低分辨率真值图像计算LPIPS loss $\mathcal{L}_{\text{LPIPS}}$，使得...(补充完善)。The total supervision is given by:
Following Scaffold-GS, we incorporate the SSIM loss $\mathcal{L}_{\text{SSIM}}$ and volume regularization loss $\mathcal{L}_{\text{vol}}$. Unlike most super-resolution methods~\cite{wang2022nerf,feng2024srgs,hu2024gaussiansr}, we avoid directly constraining high-resolution rendered images with low-resolution ground truth images through color-based losses. Such constraints would limit the model's ability to synthesize fine texture details and subtle high-frequency elements that are essential for realistic super-resolution. To enhance fidelity, we instead downsample the high-resolution rendered images and compute the LPIPS loss $\mathcal{L}_{\text{LPIPS}}$ against low-resolution ground truth images, ensuring perceptual consistency across resolutions. The total supervision is given by:
\begin{equation}
\mathcal{L} = \mathcal{L}_{\text{rec}} + \lambda_{SSIM} \mathcal{L}_{\text{SSIM}} + \lambda_{\text{vol}} \mathcal{L}_{\text{vol}} +  \lambda_{LPIPS} \mathcal{L}_{\text{LPIPS}}
\end{equation}
\section{Experiments}

\subsection{Experimental Setups}

\begin{figure*}[!ht]
\centering
\includegraphics[width=0.95\linewidth]{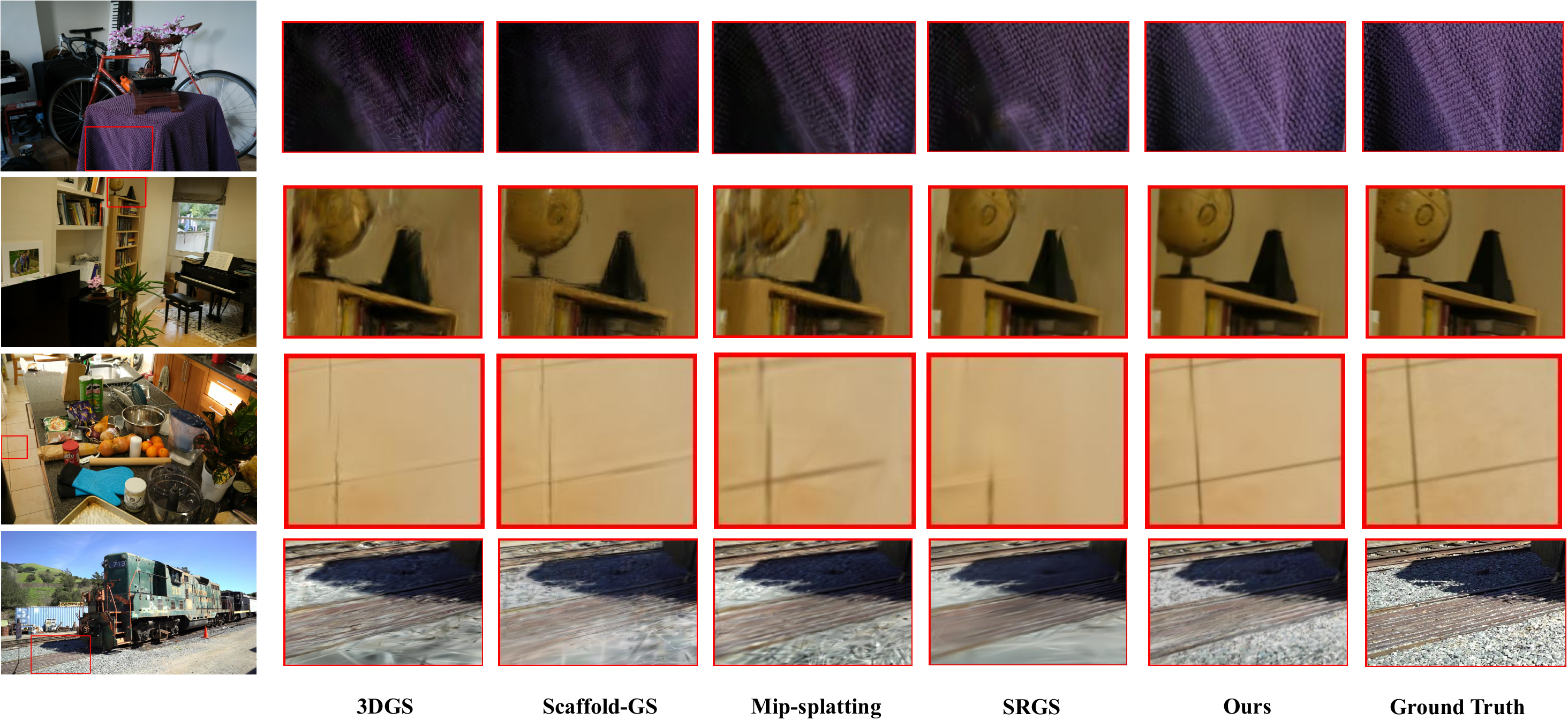}
\caption{\textit{Qualitative comparison of the HRNVS ($\times 4$) on real-world datasets.} We highlight the difference with colored patches.}
\label{fig:real_world}
\end{figure*}

\subsubsection{Datasets and metrics}

We conduct a comprehensive evaluation of our proposed model using PSNR, SSIM, LPIPS~\cite{zhang2018unreasonable} and FID metrics. For LPIPS, we use VGG~\cite{simonyan2014very} as the backbone. Our evaluation covers 21 scenes from both real-world and synthetic datasets, including 9 scenes from Mip-NeRF360~\cite{barron2022mip}, 2 scenes from Deep Blending~\cite{hedman2018deep}, 2 scenes from Tanks\&Temples~\cite{knapitsch2017tanks} and 8 scenes from LLFF~\cite{mildenhall2019local, mildenhall2021nerf}. For Mip-NeRF360 and LLFF, We downsample the training views by a factor of 8 as low-resolution inputs for the $\times 4$ HRNVS tasks. For Deep Blending and Tanks\&Temples, we downsample the training views by a factor of 4 as inputs.

\subsubsection{Baselines}

To validate the effectiveness of our method, we conduct comparisons with several prominent existing approaches. For 3DGS~\cite{kerbl20233d} and Scaffold-GS~\cite{lu2024scaffold}, we train using low-resolution input views and directly render high-resolution images. 
% Additionally, to maintain fairness in the comparisons, we utilize the same pretrained SISR model, SwinIR~\cite{liang2021swinir}, to upscale the low-resolution images rendered by 3DGS, referring to this as 3DGS-SwinIR. 
For Mip-splatting~\cite{yu2024mip}, which is a full-scale NVS method, we run the source code with its multi-resolution training setup. Regarding NeRF-SR~\cite{wang2022nerf} and GaussianSR~\cite{hu2024gaussiansr} 
%FastSR-NeRF~\cite{lin2024fastsr}, SuperGaussian~\cite{shen2025supergaussian}, 
we directly cite the results from respective papers under the same settings. And for SRGS~\cite{feng2024srgs}, we run the source code to obtain both qualitative and quantitative under identical conditions. Notably, to maintain fairness in the comparisons, we utilize the same pretrained SISR model, SwinIR~\cite{liang2021swinir} to upsample the low-resolution input views, mirroring SRGS~\cite{feng2024srgs}.

\begin{figure*}[!ht]
\centering
\includegraphics[width=0.95\linewidth]{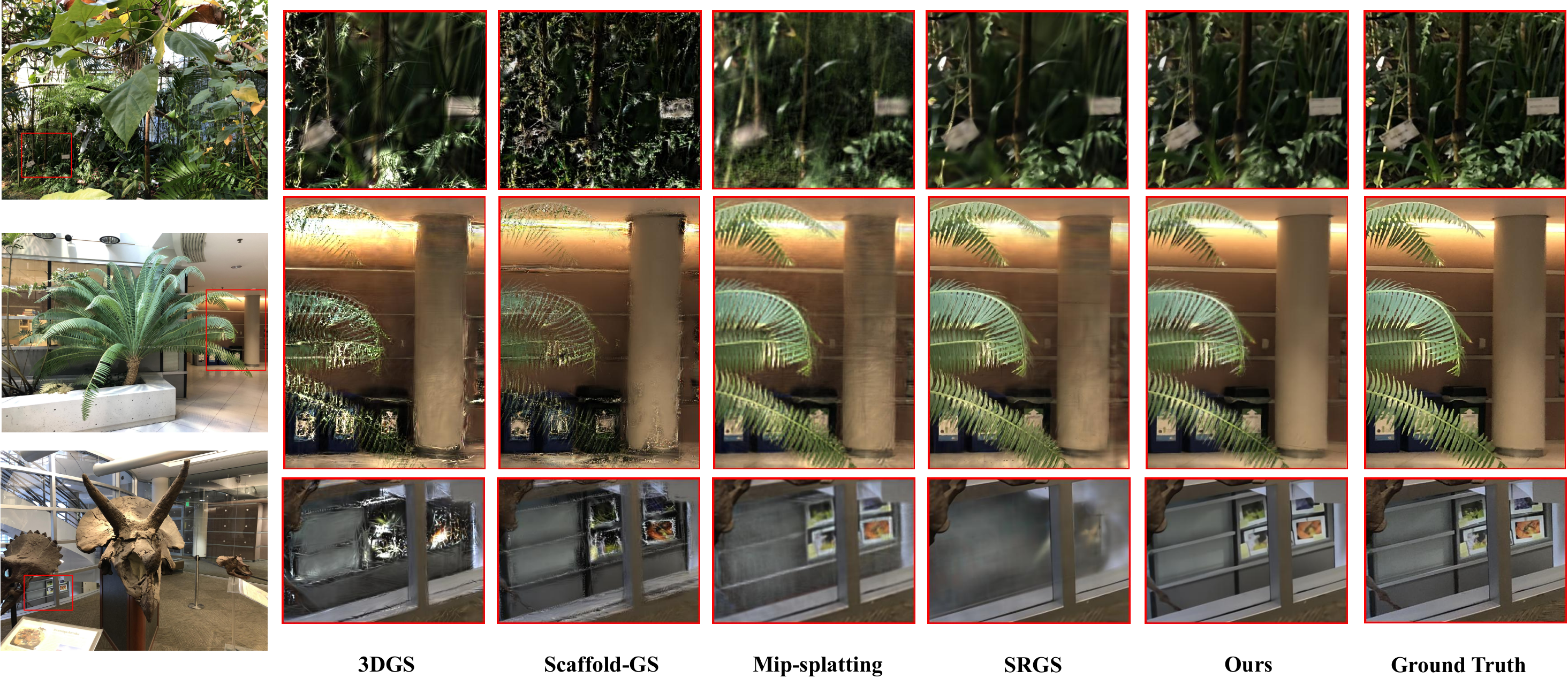}
\caption{\textit{Qualitative comparison of the HRNVS ($\times 4$) on LLFF dataset.} We highlight the difference with colored patches.}
\label{fig:llff}
\end{figure*}

\subsubsection{Implementation details}

We implement our framework based on the Scaffold-GS~\cite{kerbl20233d} source code and modify the differentiable Gaussian rasterization to include softmax-scaling depth and uncertainty rendering. %For a fair comparison, we also choose SwinIR~\cite{liang2021swinir} as the pretrained SISR model, mirroring SRGS~\cite{feng2024srgs}.
For the anchor feature, we employ a 16-dimensional feature from the latent feature field and a 16-dimensional supplementary feature bound to each anchor. In our multi-view voting densification strategy, we select pixels with loss exceeding 0.08 for reprojection, performing an anchor growth operation every 100 iterations. For anchor refinement, we iterate for 1,000 iterations following the densification process. Additionally, we train low-resolution scenes for 30,000 iterations and high-resolution scenes for 15,000 iterations. All experiments are conducted on a single RTX A6000 GPU.

\subsection{Experimental Results}

\begin{table}[!ht]
        \footnotesize
        \centering
        \caption{\textit{Quantitative comparison for HRNVS ($\times 4$) on LLFF dataset.} LLFF: $504\times378\rightarrow 2016\times1512$.}
        \label{tab:llff}
    \begin{tabular}{L{1.6cm}|C{0.8cm}C{0.8cm}C{0.8cm}C{0.8cm}} % {c|ccc|ccc|ccc}
\toprule
Dataset     & \multicolumn{4}{c}{LLFF}   \\
Method\&Metric      & PSNR$\uparrow$   & SSIM$\uparrow$  & LPIPS$\downarrow$ & FID$\downarrow$   \\ \midrule

3DGS            &17.55 &0.547 &0.448 &130.31 \\

% 3DGS-SwinIR     &28.82 &0.933 &0.073\\

% TensoRF         &28.01 &0.910 &0.113 \\

Scaffold-GS     &18.15 &0.579 &0.424 &142.60 \\

NeRF-SR         &25.13 &0.730 &- &- \\

% FastSR-NeRF     &30.47 &0.944 &0.075 \\

Mip-splatting   &23.68 &0.743 &0.332 &59.45 \\
                
% GaussianSR      &28.37 &0.924 &0.087 \\

% SuperGaussian    &28.44 &0.923 &0.067 \\
                
SRGS            &23.43 &0.770 &0.308 & 61.84 \\           
                \midrule
              
\textbf{Ours}   &\textbf{25.43} &\textbf{0.789} &\textbf{0.276} &\textbf{34.65}  \\
\bottomrule
\end{tabular}
% \vspace{-4mm}
\end{table}

\subsubsection{Quantitative comparison}

% Table~\ref{tab:real_world} and Table~\ref{tab:llff} showcase quantitative comparison results for the HRNVS tasks on the real-world and LLFF datasets. On the LLFF dataset, SuperGS achieves the best performance with a PSNR of 25.43, SSIM of 0.789, and LPIPS of 0.276, surpassing the next best competitor by +0.30dB in PSNR and +0.019 in SSIM. More importantly, SuperGS significantly reduces the FID score to 34.65, representing a remarkable 41.7\% improvement over the next best method. This substantial decrease in FID indicates that our method generates visually more coherent and perceptually accurate reconstructions. The advantages of SuperGS become even more pronounced on challenging real-world datasets. Across all three datasets, our method consistently demonstrates superior performance in terms of PSNR and SSIM metrics. Particularly on Deep Blending, SuperGS reaches 30.05 PSNR, outperforming SRGS by +0.65dB. The perceptual quality improvements are even more significant, with FID reductions of 12.11 points (26.1\%) on Mip-NeRF360 and 5.82 points (16.8\%) on Tanks\&Temples compared to the best alternative method.

Tables~\ref{tab:real_world} and~\ref{tab:llff} present quantitative comparison results for HRNVS tasks on real-world and LLFF datasets. On the LLFF dataset, SuperGS achieves superior performance, outperforming NeRF-SR by +0.30dB in PSNR. More significantly, SuperGS reduces the FID score to 34.65, a 41.7\% improvement over Mip-Splatting, indicating more visually coherent and perceptually accurate reconstructions. Across all three real-world datasets, our method consistently delivers superior PSNR and SSIM metrics. On Deep Blending, SuperGS reaches 30.05 PSNR, surpassing SRGS by +0.65dB. Perceptual quality improvements are particularly notable, with FID reductions of 12.11 points (26.1\%) on Mip-NeRF360 and 5.82 points (16.8\%) on Tanks\&Temples compared to the best alternative method.

\subsubsection{Qualitative comparison}

As shown in Figure \ref{fig:real_world} and ~\ref{fig:llff}, 3DGS and Scaffold-GS exhibits severe low-frequency artifacts due to the lack of high-resolution information, while Mip-splatting struggles with insufficient detail recovery due to missing high-frequency information. SRGS also exhibits significant artifacts and reconstruction errors. In contrast, our method demonstrates superior precision and fidelity in detail recovery while substantially reducing artifacts that severely affect the visual quality of other methods.

\subsection{Ablation Studies}

% \begin{figure}[!ht]
% \centering
% \includegraphics[width=1.0\linewidth]{figures/ablation_comparison.pdf}
% \caption{\textit{Comparison of Ablation Study.}}
% \label{fig:ablation comparison}
% % \vspace{-3mm}
% \end{figure}

We conduct ablation experiments on the LLFF dataset. Quantitative results are shown in Table~\ref{tab:ablation}.

\begin{table}[!ht]
        \footnotesize
        \centering
        \caption{\textit{Ablation study results for HRNVS ($\times 4$) on LLFF dataset.}}
        \label{tab:ablation}
% \vspace{-2mm}
\resizebox{\linewidth}{!}{
    \begin{tabular}{L{2cm}|C{0.8cm}C{0.8cm}C{0.8cm}C{0.8cm}} % {c|ccc|ccc|ccc}
\toprule
Dataset     & \multicolumn{4}{c}{LLFF}   \\
Method\&Metric      & PSNR$\uparrow$   & SSIM$\uparrow$  & LPIPS$\downarrow$ & FID$\downarrow$   \\ \midrule

w/o Feature Field            &25.28 &0.786 &\textbf{0.273} &39.08 \\

w/o Densification            &24.89 &0.768 &0.396 &47.69 \\

w/o Uncertainty         &25.35 &0.774 &0.277 &38.45 \\
              
\textbf{Ours-full}   &\textbf{25.43} &\textbf{0.789} &0.276 &\textbf{34.65}  \\
\bottomrule
\end{tabular}
}
% \vspace{-7mm}
\end{table}

\subsubsection{Effectiveness of latent feature field}

We evaluated two approaches: using only low-resolution anchor features as initialization for high-resolution learning, versus employing our feature field to provide foundational texture information for high-resolution learning. The latter approach demonstrated better performance, achieving a 0.15dB improvement in PSNR on the LLFF dataset.

\subsubsection{Effectiveness of multi-view consistent densification}

We compared our proposed multi-view voting anchor densification strategy against the gradient-based anchor growth policy from Scaffold-GS~\cite{lu2024scaffold}. Our method not only improved PSNR by 0.54dB but also required significantly fewer anchors than the gradient-based approach, as shown in Figure~\ref{fig:densify_comparison}. This reduces memory requirements while preventing overfitting to pseudo training views.

\subsubsection{Effectiveness of uncertainty-aware learning}

The uncertainty-aware augmented learning module further refines scene reconstruction and mitigates the impact of pseudo-label errors. This component provides an additional 0.08dB improvement in PSNR.

\section{Conclusion}

In this work, we present SuperGS, an expansion of Gaussian Splatting for super-resolution scene reconstruction. Our method is based on a two-stage coarse-to-fine framework, introducing a latent feature field to represent the low-resolution scene, which serves as both initialization and foundational information for high-resolution reconstruction. To address the multi-view inconsistency of pseudo-labels, we introduce multi-view voting densification, ensuring consistency and effectiveness while avoiding overfitting. Additionally, we model scene uncertainty by learning variational features, which is further used to refines scene representation and guides supervision. The results consistently demonstrate the superior performance of our method on the HRNVS task.

% \clearpage
% \newpage
\bibliographystyle{ACM-Reference-Format}
\balance
\bibliography{ref}

\end{document}